cs.CL/9809051   23 Sep 1998

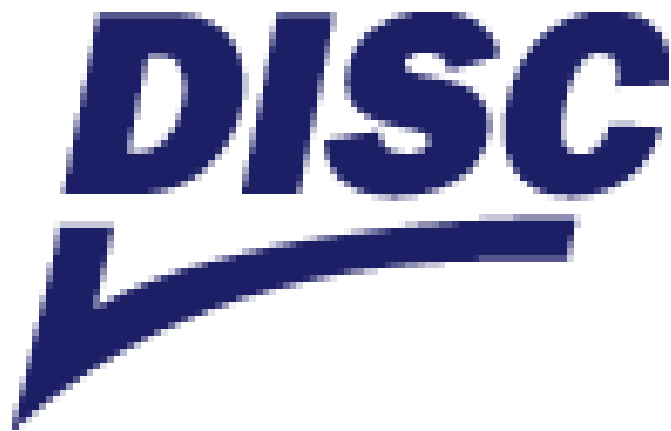

Deliverable D5.1

**Periodic Progress Report No. 1
Covering the period from: 1.6.97 to 30.5.98**

5 June 1998

Esprit Long-Term Research Concerted Action No. 24823

**Spoken Language Dialogue Systems and Components: Best practice in development and evaluation.**

# DISC

| TITLE | PPR1 Periodic Progress Report No. 1 |
| --- | --- |
| | Covering the period from 1.6.97 to 30.5.98 |
| PROJECT | DISC (Esprit Long-Term Research Concerted Action No. 24823) |
| EDITORS | Niels Ole Bernsen and Laila Dybkjær, The Maersk Institute |
| AUTHORS | The DISC Consortium |
| ISSUE DATE | 5 June 1998 |
| DOCUMENT ID | PPR1 |
| VERSION | 0.1 |
| STATUS | Draft |
| NO OF PAGES | 27 |
| WP NUMBER | 5 |
| LOCATION | DISC-PPR1-5.6.98 |
| KEYWORDS | WP5, spoken language dialogue systems, PPR |

**Document Evolution**

| Version | Date | Status | Notes |
| --- | --- | --- | --- |
| 0.1 | 05/06/98 | Draft | First draft published for comments from partners. |
| | | | |

# Spoken Language Dialogue Systems and Components: Best practice in development and evaluation

## DISC 24823
## Periodic Progress Report 1
## Basic Details of the Action

### Co-ordinator

The Maersk Institute
10, Science Park
5230 Odense
Denmark

### Partners

Daimler-Benz Research and Technology, Ulm, Germany
European Language and Speech Network (Elsnet), Utrecht, The Netherlands
Institut für Maschinelle Sprachverarbeitung, Stuttgart University, Germany
Kungliga Tekniska Högskolan (KTH), Stockholm, Sweden
Laboratoire d'Informatique pour la Mécanique et les Sciences de l'Ingénieur (LIMSI), Paris, France
Vocalis LTD, Cambridge, UK

### Co-ordinator Contact Point

Prof. Niels Ole Bernsen
Tel. (+45) 65 57 35 44
Fax (+45) 66 15 76 97
email: nob@mip.ou.dk

# Table of Contents

# 1. Executive Summary

## 1.1 Abstract

This project aims to (a) build an in-depth understanding of the state-of-the-art in spoken language dialogue systems (SLDSs) and components development and evaluation with the purpose of (b) developing a first best practice methodology in the field. The methodology will be accompanied by (c) a series of development and evaluation support tools. To the limited extent possible within the duration of the project, the draft versions of the methodology and the tools will be (d) tested by SLDS developers from industry and research, and will be (e) packaged to best suit their needs. In the first year of DISC, (a) has been accomplished, and (b) and (c) have started. A proposal to complete the work proposed above by adding 12 months to the 18 months of the present project, has been submitted to Esprit Long-Term Research in March 1998.

## 1.2 Review of Aims

The DISC aims listed in Section 1.1 above, may appear more worthwhile than ever in the light of recent developments. Spoken language dialogue systems are gaining ground in the market place as witnessed by forefront products installed in Europe by, among others, Vocalis, Daimler-Benz, Philips Aachen, Lernout and Hauspie and CSELT, and by the rapid spread of known SLDS technologies by European telecoms and others. During the first year of DISC, DARPA in the US has decided to start building up strength in the field in view of the massive build-up in speech technology in the US (cf. Business Week 15.2.1998). Starting from their pre-DISC theoretical strengths and experience in development and evaluation of SLDSs and their components, the DISC partners have spent the first year testing and revising the lessons learnt from that experience, leading to a deeper and broader understanding of the issues involved in developing and evaluating spoken language dialogue systems and their components. It is possible that the results of DISC Year 1 have given the Consortium as strong a basis as it is possible to have today for drafting a first best practice methodology in the field.

## 1.3 Progress and Results

During the first year of DISC, WP1 has been completed in draft form and work has started on WP2 (tools) and WP3 (best practice draft). WP4 (dissemination) has been active throughout and is currently "changing gear" with the arrival of the first public DISC results as well as the much larger batch of results that can be accessed by the Advisory Panel members.

For the field, DISC is highly interdisciplinary, addressing virtually all *aspects* of spoken language dialogue systems, including:

- speech recognition
- speech generation
- natural language understanding and generation
- dialogue management
- human factors
- systems integration

In addition to this "horizontal" spread of expertise, the "vertical" spread of expertise ranges from industrial development (all aspects) to theoretical research into human-human spoken dialogue.

**WP1:** *Results:*
During Year 1, all partners have been doing parallel exercises in empirical information gathering through email contacts with systems and components developers, telephone interviews, video conferencing, site visits and demonstrations as well as through analysis of scientific papers and internal documentation; systems and component analysis; current development and evaluation practice analysis; theory-informed representation of the findings made; and presentation of those findings to the developers for their comments. The common methodology adopted for this process aimed to uncover current practice, has made it possible to look for and present the findings in a uniformly structured way.

The common methodology followed in Year 1 can, moreover, be described as a combined process of top-down hypothesis testing and bottom-up empirical investigation. The *hypotheses* to be tested were the 'grid' and life-cycle models adopted at the outset (see Delieverable D1.1 and Bernsen et al. 1998). We knew in advance that these models were incomplete, based, as they were, on in-depth analysis of only a small fraction of state-of-the-art SLDSs and components. The open question was how the models would have developed, grown and changed after the first year. The *bottom-up* part of the methodology was to analyse in depth the following common exemplars with respect to (i) one or more of the aspects mentioned above and (ii) in almost all cases from a dual grid-and-life cycle point of view:

*Daimler-Benz Dialogue Manager*
Aspects analysed: dialogue management.

*Daimler-Benz Parser*
Aspects analysed: natural language understanding and generation.

*Danish Dialogue System*
Aspects analysed: dialogue management, human factors.

*French Arise*
Aspects analysed: speech recognition, speech generation, natural language understanding and generation, dialogue management.

*Operetta*
Aspects analysed: speech recognition, human factors, systems integration.

*Vad/SpeechTel*
Aspects analysed: speech recognition, systems integration.

*Verbmobil*
Aspects analysed: speech recognition, speech generation, natural language understanding and generation, dialogue management, human factors, systems integration.

*Waxholm*
Aspects analysed: speech generation, natural language understanding and generation, dialogue management, human factors, systems integration.

In two cases, i.e. dialogue management in Verbmobil and Waxholm, two different teams investigated the same aspect of a system or component. This was done to test the assumption that, due to the lack of anything resembling a unifying theory, dialogue management in particular would raise theoretical issues whose solution would benefit from the combined efforts of several DISC partners. The assumption was confirmed and the theoretical discussions occasioned by the parallel analyses of Verbmobil and Waxholm have proved eminently worthwhile.

In total, 26 SLDSs and components analyses were carried out, leading to approx. 50 grid and life cycle representations in internal and confidential Working Papers.

Based on the exemplars analysis, six synthesis Working Papers were produced, one per aspect investigated (see Deliverables 1.2, 1.3, 1.4, 1.5, 1.6 and 1.7, respectively). Finally, based on the synthesis papers, a draft "super" synthesis paper was produced (see D1.8) whose general conclusions will be discussed and, most likely, significantly augmented at the Year 1 Workshop. D1.8 also describes how the work during Year 1 of DISC was carried out in practice, obstacles met with, etc. As an appendix to D1.8, the "revised hypotheses" concerning the contents of (comparatively more) adequate grid and life cycle models relative to those adopted at the start of DISC, are presented in their "empty" forms.

*Delays*
A first version of D.1.8., serving as a reading guide to the WP-1 deliverables, was produced in May 1998. The fact that all aspects papers (synthesis reports D-1.2 through D-1.7) have been delayed due to the general start-up delay of the project, has implied that the section on trends in development and evaluation is so far rather brief and sketchy. It will be expanded to take into consideration the conclusions of all the aspects synthesis reports.This will be done in the second half of June and in early July 1998.
A second version of D.1.8. will be produced by 10.7.98.

**WP2:** *Progress*
In parallel with development of the DISC draft best practice methodology, work is underway to develop the following concepts and software tools for testing in industrial and academic environments:

**LIMSI**: Survey of existing and easily available platforms and development methods for testing and enhancing the performance of speech recognition components.
*Progress made:* This report was added to ensure that Disc takes into account existing methodologies used by the speech recognition community, and serves as a basis for D2.2. See also D2.1.
Guidelines and testing protocols for the development of speech recognition components for SLDSs.
*Progress made:* About half the work on D2.2 is finished. D2.2 will be delivered by the end of month 14 (July 98).

**KTH**: A survey of existing methods and tools for the development and evaluation of speech synthesis and speech synthesis quality in SLDSs.
*Progress made:* Available tools for producing speech synthesis have been identified and investigated. These include some freeware for speech synthesiser software, the possibility of producing one's own concatenative synthesis diphone database, and text-to-speech systems. These are available for free for non-commercial and non-military proposes. A first draft description of synthesis tools exists.
Software tool for evaluation of speech synthesis components in SLDSs.
*Progress made:* We are investigating evaluation methods for speech synthesis with the intention to construct a protocol for speech synthesis evaluation that can be useful in different situations, especially for spoken language dialogue systems.
There has been some delay due to lack of staff, 2 person months.

**IMS**   Survey of tools and methods for the acquisition of lexical data for SLDSs. Guidelines and tool specifications for checking, enhancing and extending the lexical coverage of SLDSs.

Guidelines for the construction of linguistic resources for SLDSs. Guidelines for the representation of the relevant types of information.

*Progress made:* Both deliverables are being constructed in parallel. Part of the guidelines will be derived from the detailed analysis of the WP-1 outcome. Another part will be based on the experience gained in ongoing work in collaboration with DISC partner Daimler-Benz, on the development of tools for ensuring consistency in the linking of internal representations and representations of the Daimler-Benz task modules. This work is being done in the framework of a diploma thesis (Thomas Witzemann, of IMS, currently at Daimler-Benz Research centre, Ulm). A draft outline for discussion with other DISC partners will be made available in the course of June 1998. We expect a small delay in the preparation of the final version of the deliverable.

**MIP**   State-of-the-art survey of existing dialogue management and human factors tools.

*Progress made:* A DISC Working Paper is in progress.

Concepts and a diagnostic methodology for the identification of user-system interaction problems, their typology, severity and remedies. Software tool in support of cooperative system dialogue design.

*Progress made:* A DISC Working Paper is in progress. Seven publications are: Bernsen, Dybkjær and Dybkjær 1997 (IEEE Computer), Bernsen, Dybkjær and Dybkjær 1997 (book chapter), Bernsen, Dybkjær, Dybkjær and Zinkevicius 1997 (Eurospeech), Bernsen, Dybkjær and Dybkjær 1998 (book), Dybkjær, Bernsen, and Dybkjær 1997 (ACL Workshop), Dybkjær, Bernsen, and Dybkjær 1997 (book chapter), Dybkjær, Bernsen, and Dybkjær 1998 (International Journal of Human Computer Studies/Knowledge Acquisition, to appear).

Software tool in support of speech functionality decisions in early design.

*Progress made:* A DISC Working Paper is in progress. Three publications are: Bernsen 1997 (Speech Communication), Bernsen, Dybkjær and Dybkjær 1998 (book) and Bernsen and Dybkjær 1998 (ICSLP submission).

We expect 1-2 months delay in the preparation of the final versions of the MIP WP2 deliverables.

**WP3:** *Progress*

A first draft of a best practise methodology has been composed. This draft draws on the deliverables of WP1. Later, information gathered in WP2 will be added to this document and it will serve as a draft for tests in WP3. See also D3.1.

**WP4:** *Progress*

A web site was set up, with three different compartments and audiences:

- a compartment for internal use, where working documents and other relevant information is stored, including an email archive (http://www.elsnet.org/disc-internal)
- a public compartment where SLDs and the DISC results will be promoted; at this moment there is no DISC output available yet, and the page contains some general information on the project and a list of publicly accessible SLD systems (http://www.elsnet.org/disc)
- a compartment used for information exchange with the DISC Advisory Panel (cf below) (http://www.elsnet.org/disc/ap)

In collaboration with the DISC partners, DISC-relevant information was, and is being, collected and stored on the DISC internal web pages, and in part made available via the public pages. The DISC www site made a slow start due to under-staffing of the ELSNET office (Utrecht and Edinburgh) between November and February.

Flyers have been produced, and distributed by mail, at various conferences and other events.

An Advisory Panel of experts has been created, currently with 23 members from academia (6) and industry (14), from 13 countries. As of early May 1998, the composition by country is as follows:
- 6   Germany
- 2   USA
- 2   UK
- 2   France
- 2   Denmark
- 1   The Netherlands
- 1   Switzerland
- 1   Sweden
- 1   Russia
- 1   Romania
- 1   Japan
- 1   Italy
- 1   Ireland
- 1   Finland

The results of the first year of DISC have been made available to the members of the Advisory Panel, and will be the topic of a workshop with DISC participants and Panel members on June 8 and 9, 1998, in Stuttgart.

**WP5:** *Progress*

WP5 is the management Work Package. The following items are reported here:

(1) During the first DISC workshop it was decided to change the planned Industrial Advisory Panel (IAP) into an Advisory Panel (AP) which could have both academic and industrial members, and which was envisaged to become considerably larger than the planned IAP. The reasons for the change were that many research groups continue to be involved in prototyping SLDSs and components and hence could be assumed to provide important input for DISC; and that it is desirable to involve as many stakeholders in the discussion of the emerging DISC best practice methodology as possible.

(2) During the first year of DISC, Linköping University has kindly stepped in as a subcontractor to carry out KTH's part of the work on dialogue management.

(3) Due to lack of person-power, KTH wanted another project partner to take over one person month on system integration in WP1. LIMSI carried out this work. However, since LIMSI could not put in additional person-power, the problem was solved by having IMS take over one person month of work on the natural language understanding and generation aspect from LIMSI. In brief, this means that one person month was transferred from KTH to IMS. The formalities are still waiting to be resolved.

(4) Most WP1 deliverables have had a delay of a month to a month and a half (see Section 4.2). This has in all cases been due to the complex logistics involved in obtaining full information on exemplars and verifying the analyses produced with the developers. However, we have now caught up with these delays and the project as a whole is on schedule for the work to be done in Year 2.

(5) One more workshop than originally planned was held at Eurospeech in Rhodes in September 1997. This workshop helped streamline the DISC methodology at an important point in the project.

(6) The WP2 deliverables ordering has been slightly changed in order to include a current practice overview. Thus, the WP2 deliverables now include state-of-the-art surveys with respect to existing tools and methods for the aspects analysed in WP1 except systems integration. The revised WP2 description is as follows (cf. Section 3.1):

**Tasks.** WP2 includes the following tasks:

**T2.1-T2.4: Development and test of concepts and software tools**

The following four tasks concern the development of best practice concepts and software tools. WP2 does not envision the development of concepts and software tools related to system integration. However, each partner will contribute to the integration aspects by including a description and evaluation of the platforms and tools that are easily available (mainly, but not restricted to, public domain and academic distribution) and relevant to the particular SLDS aspect they are dealing with in DISC.

The planned efforts focus on the aspects addressed below. The development work will be planned in detail at W1, reviewed at W2 and W3 and presented at W4. Development and test cases will be selected at W1. Cases from ongoing projects for use in T2.1-T2.4 are available at the DISC partner sites.

**Task T2.1: Speech recognition.**

*Contributor and effort:* LIMSI 6, IMS 1.

*Description:* Survey existing platforms and development methods for testing, and enhancing the performance of Speech Recognition components. To document and exemplify the current best practices, this survey will only concern easily available and well known platforms (i.e. mainly, but not restricted to, public domain and academic distribution), as well as those in use at the DISC partner sites (when information disclosure does not become an issue). When possible, DISC-relevant relationships will be described between issues specific to Speech Recognition and issues specific to Spoken Language Dialogue Systems integration. To clearly distinguish T2.1 work from work in WP1, the focus of the T2.1 work will be "support" of best practice (development environment and scoring).

Develop concepts and a methodology for best practice in building the speech recognition components (acoustic models, lexicon, language models) for SLDS. The methodology includes definition of task-specific vocabulary as well as the needs for data collection (amounts and type of accoustic and textual data) for training the acoustic and language models. These actions will be coordinated with T2.3 (a) and (b). Assess dependence of system performance on training data. The methodology must account for constraints imposed by the application (real-time processing, acoustic environment and background noise or channel conditions, signal capture, task domain, user population etc.). Recognition development is typically an interative process involving evaluation and modification. Guidelines for development will be elaborated, along with testing protocols. Techniques to cope with spoken language specificities such as hesitations, false starts, repetitions, pauses, variable speaking rates found during user interaction with SLDS will be specified.

*Duration:* Month 1-13.

**T2.2: Speech synthesis.**

*Contributors and effort:* KTH: 6.

*Description:* (i) A survey of existing methods and tools for development and evaluation of speech synthesis and speech synthesis quality in SLDSs. This should take into account that there has over the last few years been a shift from formant synthesis to waveform (concatenation) synthesis and should look at the methods and tools with this in mind. (ii) Develop a test protocol for speech synthesis that can be transferred to industry and implement this as a software tool. The tool will enable developers to evaluate different variants of speech synthesis and their usefulness to the particular system they are building. The tool will consist of a protocol containing a number of questions that are relevant in choosing between different speech synthesis systems for a particular SLDS system to be developed.

*Duration:* Month 1-13.

### T2.3: Language understanding and generation.

*Contributors and effort:* IMS: 6, LIMSI: 1.

*Description:* Survey and develop methods and tools for testing, enhancing and possibly extending the linguistic (in particular: lexical) coverage of SLDSs at the levels of morphology/morphosyntax and syntax. This includes (a) a survey of existing tools for lexical (and, to some extent, grammatical) acquisition for from corpora SLDSs; (b) the development of guidelines for the construction of linguistic resources for SLDSs, with a view to their partitioning (core vs. extensions; sublanguage, etc.), their extensibility, ease of maintainance, efficiency and robustness; (c) the development of guidelines and specifications for tools to be used for checking and enhancing the lexical coverage of SLDSs, in particular with respect to spoken language/dialogue phenomena. Depending on the approach (integrated resources for SR and NL tasks vs. separation of different types of knowledge), different versions of the diagnosis and enhancement tools need to be specified. The guidelines will include proposals for the representation of the relevant types of information, for each approach.

Actions (b) and (c) will be carried out in close coordination with Task 2.1.

*Duration:* Month 1-13.

### T2.4: Dialogue management and human factors.

*Contributors and effort:* MIP: 9.

*Description:* (i) Develop concepts and a rigorous diagnostic methodology for the identification and quantification of user-system interaction problems, their typology, severity and remedies. The methodology will enable developers to follow strict procedures throughout their diagnostic analysis of test data. (ii) Test a body of candidate practical guidelines for cooperative system dialogue design for transfer to industry, and implement these in a software tool. The tool will enable developers to detect and repair dialogue design problems prior to implementation. (iii) Develop a software tool in support of speech functionality decisions in early design. The tool will help developers decide, during early specification, whether speech modalities should be included in the application to be developed, which modalities should be included and under which conditions. (iv) Perform a state-of-the-art survey of existing dialogue management and human factors tools.

*Duration:* Month 1-13.

## 1.4 Internal Collaboration

In order to support cross-site collaboration and consensus-building, DISC has been designed as a highly collaborative project with at least two different DISC partners being involved in analysing each aspect of SLDSs and their components. Partly because of this design, but also in response to needs arising as the project unfolded, internal interaction has been really lively during the first year of DISC. Three workshops have been held so far in Stuttgart (three days), Rhodes (one half day during Eurospeech 1997), and Ulm (three days), respectively. In addition, four other collaborative mechanisms have been used. The second mechanism was the many cross-site visits which had to be undertaken in order to interview and discuss with the developers, attend demonstrations, etc. Summarising, the following 14 cross-site visits have been made, listing the venue followed by the sites from whence the visitors came:

Cambridge: KTH, LIMSI.
Paris: Linköping.

Odense: Linköping.
Saarbrücken: IMS, LIMSI, MIP, Vocalis.
Stockholm: IMS, LIMSI, MIP, Vocalis.
Ulm: IMS, MIP using video conferencing.

The third collaborative mechanism was the jointly authored synthesis papers per aspect (see Section 1.3) as well as the "super synthesis" paper, which all have cross-site authorship, as follows:

D1.2: LIMSI, KTH.
D1.3: KTH, LIMSI.
D1.4: IMS, LIMSI.
D1.5: MIP, IMS, Linköping.
D1.6: Vocalis, MIP.
D1.7: Vocalis, LIMSI.
D1.8: IMS, MIP.

The cross-site authorship frequently turned out to force theoretical discussions which might not have happened otherwise within the project. These are still going on.

The fourth mechanism of collaboration was the writing of joint publications. At this point, one joint DISC publication has been produced - for the Granada Conference in May 1998. At the Year 1 Workshop, the DISC partners will discuss how to publish the results of the first year of DISC.

A fifth mechanism of collaboration has been the joint DISC website established by Elsnet, which has been used extensively for uploading draft working papers for easy access by the partners. Each uploading of a paper onto the "Partners only" web pages was accompanied by a message from the website robot to all DISC partners that a paper with a specified identity was now available on the website.

## 1.5 External Collaboration and Use of Results

For the purpose of this section, "external collaboration" includes any DISC-related collaboration undertaken by a DISC partner with non-DISC partners, including site-internal collaboration.

**MIP**: the emerging DISC best practice methodology is being brought to the test through development of three different SLDSs: one in-house for user interaction with two collaborating robots aimed to demonstrate latest advances in production technology; one for integrating interactive speech-based web search into a European research project; and one for a prototype SLDS in collaboration with local service industry. The research goal behind those three SLDSs is to investigate generic dialogue management through building a dialogue manager which can handle three very different applications.
In addition, DISC results are being used in the following projects:
ELSE project.
MATE project.
*Talks and seminars:*
SALT workshop 1997: Presentation of DISC given by L. Dybkjær.
ACL/EACL 1997: Presentation of MIP WP2 related results given by L. Dybkjær.
Eurospeech 1997: Presentation of MIP WP2 related results given by L. Dybkjær.

**LIMSI:** DISC results are being used in the following projects:
ELSE project
Aupelf B1 and B2 activities
LE ARISE project
DARPA Communicator (preliminary contacts)
*Talks and seminars:*
LREC'98: Keynote L. Lamel 30/05/98 Granada: On the Role of Evaluation in Spoken Language System Development.
LREC'98: Disc talk (for the paper on the DISC approach [Dybkjær et al. 1998]) given by L. Lamel 28/05/98.

**KTH:**
-

**IMS:**
Site-internal collaborations using DISC results
- SFB 340 (Theoretical Foundations for Computational Linguistics): DISC survey results used in the work on the semantic and pragmatic modelling of dialogues (Jan v. Kuppevelt)
- Verbmobil, Workpackage on Transfer: DISC results (especially from DISC analysis of Natural Language and of Dialogue Managament aspects) used in work on the formulation of lexical and (dialogue-context-dependent) grammar rules for transfer; similarly: the Verbmobil system implementation available through STRs involvement in the Verbmobil project has been used in the DISC WP-1 work.
- MATE (LE-4): The DISC survey results are used as a background for the definition of requirements with respect to the functionalities required in a multilevel-annotated corpus: SLDS developers are seen as a major client of the type of corpora to be produced in MATE (Ulrich Heid, Andreas Mengel (MATE collaborator in STR)); Experience from the presence of explicitly modeled interactions between different levels of description in dialogues (esp. as visible from DISC WP-1 analyses of Verbmobil and Waxholm Dialogue Management) is used in MATE as an exemplification of the types of interactions between levels necessary from the point of view of SLDSs (Jan v. Kuppevelt); this also applies to cases where a thorough description of the facts would require such interaction to be explicitly modeled, but it isn't in the exemplars analysed in DISC.
*Talks and seminars:*
Heid, Ulrich: Objectives and working procedure of DISC, WP-1, at the DISC/Verbmobil joint working meeting in Saarbruecken, 28-11-1997
Seminars attended:
Heid, Ulrich: Session on Spoken Language Dialog System Evaluation (1st of 2), in the framework of the 1st Intl. Conference on Linguistic Resources and Evaluation, Granada, 28-5-1998

**Vocalis:** 'Guidelines for Advanced Voice Dialogues' Project (ESRC Project ref: L127251012 by the Department of Sociology, University of Surrey, UK and by Vocalis Ltd.).
This project is concerned with establishing guidelines for the development of advanced voice dialogues.
The work of the Guidelines project focuses on a traditional framework for dialogue design. 'Hanging off' this framework are a variety of guidelines. The guidelines are based on:
- Linear and iterative approaches
- Empirical investigation of *de facto* standards in commercial dialogues.

Central to the life-cycle is the applicability and uptake of the guidelines within an industrial context. To this end, the project began with an investigation of the current design practice in commercial environments (See Cheepen, 1996). With this as a starting point, the work has focused on identifying the different audiences within the company which are involved in dialogue design, i.e. developers, marketing/sales and researchers. Each group has a stake in the system but very different requirements in terms of the information they require in order to follow a suitable design process. Only by addressing each group in their own terms, can a process description hope to succeed.

In addition to the iteration of the process within a commercial setting, the project also challenges *de facto* standards within the industry. This includes:

- The use of human-like tokens in (English) system output prompts, e.g. 'please', 'thank you', 'I', 'you' (See Williams and Cheepen, 1998).
- The use of only verbal aural output in dialogue.

References:

Cheepen, C. (1996) "Designing advanced voice dialogues - what do designers do and what does this mean for the future?", *http://www.soc.surrey.ac.uk/research/reports*

Williams, D. M. L., C. Cheepen (1998) "Just Speak Naturally: Designing for Naturalness in Automated Spoken Dialogues", In: Proceedings of ACM SIGCHI'98, Los Angeles.

*Talks and seminars:*

Klaus Failenschmid attended Eurospeech Conference in Rhodes (22-25 September 1997).

Klaus Failenschmid attended COST Workshop (Speech Technology in the Public Telephone Network: Where are we today?), Rhodes (26-27 September 1997).

Klaus Failenschmid gave Presentation of DISC-relevant issues (Title: Spoken Language Dialogue Systems - Dreams and Reality) at Jetai '97: New Ways of Communicating in Glasgow (12-15 November 1997).

## 1.6 Information Dissemination

As shown in the list of 23 DISC-related scientific papers and books in Section 5, the DISC partners have been active participants in international research on SLDSs and components during the first year of the project. Four of those publications specifically present the DISC agenda (Dybkjær and Bernsen 1997, Elsnews, Bernsen and Dybkjær 1997, ELRA Newsletter, Bernsen and Dybkjær 1997, SALT, Bernsen and Dybkjær 1997, DISC Flyer). Given the internal and confidential character of the 50 or so working papers produced in the first year of DISC, DISC has only recently become ready to "go public" with its results based on the publicly available synthesis Working Papers described in Section 1.3. A first publication is Dybkjær et al. 1998 (Granada).

The DISC website has provided public information on DISC throughout Year 1.

Finally, the DISC website has recently been augmented with a special section for the Advisory Panel Members. This section will be used extensively during the second year of DISC as a forum for discussion and exchange of information between the DISC partners and the AP members.

There are by now 26 Advisory Panel Members. The Advisory Panel will contribute to DISC's work on SLDSs best practice by commenting on intermediate results, providing access to products, SLDS prototypes or components under development, and/or making us aware of practices, theories and tools in current use.

# 2. Update of Worldwide State-of-the-Art

## 2.1 Speech Recognition

The state-of-the-art in speech recognition for spoken language dialogue systems consists of real-time, speaker-independent, continuous speech systems with mid-size vocabularies (up to several thousand words). Most of these systems make use of statistical models of speech production. Acoustic models are typically continuous density Hidden Markov Models (HMMs) with Gaussian mixture, of phones in context. Different techniques are used to select contexts, such as decision trees, frequency of occurrence in the training data, clustering or generalized smoothing. Language models for the recognizer are ususally N-gram or class N-grams, where the statistics for these models are estimated on the language model training material. Search is based on frame synchronous dynamic programming, and beam-search pruning techniques are used to reduce the search space. Multiple decoding passes may be carried out so as to allow more accurate models to be used in later passes thus improving performance while minimalizing the additional computation time.

Most speech recognizers are written in C or C++ and can run on standard platforms without special hardware. Only signal capture and communication are particularly device dependent.

References:
"Let's Talk", in Business Week, Feb 23, 1998.
Gibbon Dafydd, Roger Moore and Richard Winski (1997). Handbook of standards and resources for spoken language systems. Mouton de Gruyter. Berlin, New York. 1997.
Steve Young, "A Review of Large-Vocabulary Continuous-Speech Recogntion," IEEE Signal Processing Magazine, Sept 1996, pp. 45-57.

## 2.2 Speech Generation

Many different types of speech synthesis systems exist today and are commercial products. They range in complexity from waveform coded spoken messages to full text-to-speech systems containing many languages. The choice between different systems will be decided by the intended use. Coded speech gives a natural speech quality while the number of messages is very restricted. Text-to-speech systems are much more flexible, but the quality of the speech varies between systems. Today, most systems demonstrate a high degree of comprehension while the quality is often perceived as unnatural.

Speech synthesis systems can normally be split up into two parts. One part contains text-to-speech rules and the other part is the speech synthesiser that produces the sound. Two main methods of speech synthesisers exist, concatenative synthesis where shorter or longer speech segments are concatenated according to rules, and formant synthesis where a speech production model consisting of sources and filters produce the speech. Speech synthesis is sometimes combined with a talking head where the mouth movements are synchronised with the acoustic speech signal to enhance comprehension. A third synthesis method, articulatory synthesis, is Text-to-speech rules decide the pronunciation of the speech both on the segment and on the sentence level. The text-to-speech rules used to be derived from linguistic knowledge about the language in question but in later years statistical methods have come into use for rule generation. Likewise, with the growing size of computer memory and faster computers more of the word pronunciation is derived from lexicon transcripts.

Speech synthesis references
http://www.speech.kth.se/info/ext_speech.html/
http://www.speech.cs.cmu.edu/comp.speech/FAQ.Packages.html/

Carlson R., Granström B., Speech Synthesis, in W Hardcastle & J Laver (editors) THE HANDBOOK OF PHONETIC SCIENCES, Blackwell Publishers Ltd, Oxford 1997, pp.768-788.

Dutoit, T.: An Introduction to Text-To-Speech Synthesis, Kluwer Academic Publishers, Dordrecht, 1997.

van Santen, J., Sproat, R., Olive, J., Hirschberg, J. (Eds.) Progress in Speech Synthesis, Springer, New York, 1997.

Syrdal, A., R. Bennett, S. Greenspan (Eds.) Applied Speech Technology, CRC Press, Boca Raton, 1995.

## 2.3 Natural Language Understanding and Generation

We are not aware of a major innovation in the field of the development and technical options for NL components in SLDSs. However, we became aware of additional evidence for the field of NL component evaluation.

Polifroni et al. 1998 describe evaluation methods for the JUPITER system, a weather forecast SLDS for Northern America. The evaluation of the NL understanding component is very similar to the "concept accuracy measure" used in the evaluation of the DaimlerBenz SLDS. JUPITER produces semantic frame representations of user queries. Evaluation of parser coverage is simply done by counting the sentences from both user queries and weather report sources (used for knowledge acquisition in JUPITER) which are analysable.

Evaluation of NL understanding is based on semantic representations and on a comparison of representations generated for new sentences with known-to-be-correct representations of analogous sentences from previous sessions. Insertions, deletions and substitutions are flagged, as in the work of DaimlerBenz (see summary report D-1.4), and counted. The procedures are automatic and use a continuous flow of user queries. Generation quality can only be assessed on the basis of human judgement.

References:

Joseph Polifroni, Stephanie Seneff, James Glass and Timothy J. Hazen: Evaluation Methodology for a Telephone-Based Conversational System', in: Proceedings of the 1st International Conference on Language Resources and Evaluation, Granada, 28-30/5/1998: 43-49.

## 2.4 Dialogue Management

Dialogue managers have so far typically been designed as part of an entire spoken language dialogue system (SLDS). Probably for this reason focus has rarely been on dialogue management for its own purpose rather than on dialogue management as one of several components which have to work together to form an SLDS. In some cases focus has actually not been on dialogue management at all. The dialogue manager was built only because it is a necessary part of an SLDS and hence a necessary evil if one wants to study, e.g., speech recognition in an SLDS environment. However, there are now ongoing efforts world-wide to develop dialogue managers which are more general-purpose and which may fairly easily be adapted to a new task domain, such as the Daimler-Benz dialogue manager.

There are not many tools available in support of building dialogue managers. A few dedicated languages exist. Common to most of these languages is that they are event-based and have a range of primitive operations that support the speech and language layers. It remains, however, an open question to which extent primitives of any of the languages just mentioned, scale up beyond relatively simple, well-structured tasks. If the language primitives needed are not provided by a specialised dialogue specification language, it may often be preferable to use a general programming language, such as Lisp, Prolog or Java, which usually also means that the resulting software will be easier to port.

## 2.5 Human Factors

Human factors cover all aspects of the interactive system design which are related to the end-user's abilities (perceptual, cognitive and motor), experience (system specific, domain specific and common sense), goals (both interactional[1] and transactional[2]) and organisational/cultural context (del Gado and Neilson, 1996). Whilst the remit of this field is broad, the theoretical and practical work tends to occupy a variety of small niches with few unifying approaches defining interactive system design on all of the dimensions noted.

The analysis with respect to Human Factors used a number of exemplars systems to evaluate current design practice against a best practice framework These were Vocalis Operetta, Waxholm and Verbmobil.

As well as specific areas of research and practical experience, the evaluation also addressed the unique requirements for the design life-cycle of interactive systems. This includes prototyping and descriptive methodologies.

This current practice evaluation has provided an overview of the Human Factors-related aspects of commercial and research spoken-dialogue systems. Those areas of particular interest have been elucidated and placed in the context of an engineering life-cycle for interactive system development. In order to provide an indication of the state-of-the-art, both individual aspects and life-cycle best practice were used to evaluate the exemplars. From this analysis the following deficiencies were identified in current practice:

*Documentation*
- There was little formal documentation at any stage of the design and implementation process except for Danish Dialogue System.
- 

*Ethnology*
- Little consideration given to organisational effects of systems and how to design for these.
- Few real users were used in evaluations.
- Context analysis was limited. The reason given for this in Waxholm and Danish Dialogue system was that the goal of the project was to research some aspect of the system ,e.g. speech recognition., rather than meet user needs. The Danish Dialogue System was a little bit different in that it focused on Human Factors. However, since the systems carried out evaluations with real subjects, it seems possible that results for the component under study will be confounded by the negative impact of a system ill-fitted to its context.

*Help*
- In NLP systems there was little explicit help capability. This could be in the form of example dialogue flows or typical utterances.

This work has provided a good understanding of current practice with respect to Human Factors in SLDSs, both research systems and commercial systems. The next step is to characterise how these deficiencies can be overcome, and provide a comprehensive description of a realistic best practice for future system developments. These will be the next stages of the DISC project (WP3).

## 2.6 System Integration

The exemplars evaluated in this Workpackage were: Verbmobil, Waxholm, Vocalis Operetta and Vocalis VAD. Evaluations were carried out by Vocalis and LIMSI.

---

[1] Related to maintaining the relationship between communicating parties. This includes ritualistic communication such as politeness.

[2] Related to the external goal of a communication, e.g. finding directions to the theatre.

A common feature of SLDSs is that they are highly complex systems which make use of a number of more or less distinct functional modules (e.g. speech recogniser, language understanding, speech synthesiser) to achieve the overall goal of providing a service or information. Depending on the application and task domain these functional modules require very different contextual information (e.g. acoustic speech models, dialogue models) and different levels of 'intelligence' (e.g. user models, dialogue history). The complexity of SLDSs stems from the need of exchanging and sharing large amounts of data between functional modules. Not only is it difficult to clearly distinguish tasks for different modules, but also is it difficult to specify the exact flow of data between two or more modules.

In currently available systems, module tasks and data exchanged between modules are defined by the context of the application. SLDS-internal systems integration of the various functional blocks is done such that the complete SLDS performs well in the task and application domain specified. Unfortunately, extending, modifying or adapting such a system can be extremely time and labour extensive. Also, the development of a new SLDS for a different task domain using pre-existing functional modules often involves re-engineering the fundamental concepts that were defined when the initial system was developed.

Results of the current practice evaluation of this aspect show that although SLDSs are tightly integrated software systems with numerous (semi-) autonomous functional modules, they tend to make use of proprietary standards and protocols. This makes modification and adaptation of the systems to a new target domain time and cost extensive. Furthermore, the systems integration life-cycles for research systems differ from the ones for commercial systems. The individual stages in the life-cycle are identical for the two types of systems, however systems integration for research systems tends to be driven by the need for integration of existing functional modules. By contrast, systems integration for commercial system tends to be driven by the need of achieving certain functionality as described by the client.

The evaluation and analysis on this aspect has benefited from the grid and life cycle work on other aspects. Systems integration is very much concerned with the whole SLDS from an 'information-exchange' point of view. Therefore, the analysis of relationships between different modules which has been added to the analysis of each individual aspect has provided valuable input to the systems integration evaluation. The importance of this has been recognised at the second workshop in Ulm.

During the current practice evaluation it has emerged that not only is it important to focus on 'SLDS-internal' communication, but also to concentrate on the way in which an SLDS is integrated with other systems such as the telephone system or LAN/WANs. This aspect is an extremely important one especially for commercial systems.

Progress in this Workpackage has been somewhat slower than expected due to the departure of a researcher (David Williams) at Vocalis. A comprehensive document describing current practice in systems integration has been produced, however, the accuracy of the information in this document has only be verified by the developers of the Vocalis Operetta and VAD for SPEECHtel. The correctness of the Waxholm and the Verbmobil information still needs be verified.

# 3. Deliverables Overview

## 3.1 Deliverable Summary Sheet

**Milestone M1 (concluding month 12):**

| Delive-rable | Task | Responsible | Due mth. | Acc. | Description |
|---|---|---|---|---|---|
| D4.1 | T4.3 | Elsnet | 1 | R | Proposal for Industrial Advisory Panel in collaboration with the DISC Management Board. **Done.** |
| D4.2 | T4.1 | Elsnet | 1 | P | DISC email list and DISC WWW pages established. Elsnet dissemination plan. **Done.** |
| D4.3 | T4.2 | Elsnet | 1, 4, 12, 18 | R/R/P/P | Four workshops. **4 workshops done in Year 1.** |
| D1.1 | T1.1 | IMS | 3 | P | Report describing the first DISC dialogue engineering best practice model. **Done.** |
| D1.2 | T1.2 | LIMSI | 10 | R | Working paper on speech recognition current practice. **Done.** |
| D1.3 | T1.3 | KTH | 10 | R | Working paper on speech generation current practice. **Done.** |
| D1.4 | T1.4 | IMS | 10 | R | Working paper on language understanding and generation current practice. **Done.** |
| D1.5 | T1.5 | MIP | 10 | R | Working paper on dialogue management current practice. **Done.** |
| D1.6 | T1.6 | Vocalis | 10 | R | Working paper on human factors current practice. **Done.** |
| D1.7 | T1.7 | Vocalis | 10 | R | Working paper on systems integration current practice. **Done.** |
| MD1.8 | T1.8 | IMS | 11 | P | Integrated report on current development and evaluation practice of SLDSs and components and its deficiencies based on the results of T1.1-T1.7 and W2. **Done.** |
| D2.1 | T2.1 | LIMSI | 8 | P | Survey of existing and easily available platforms and development methods for testing and enhancing the performance of Speech Recognition components. **Done.** |
| D3.1 | T3.1 | KTH | 12 | R | Working document on the detailed DISC best practice methodology. **Done.** |
| D5.1 | WP5 | MIP | 12 | P, A | Annual progress report. **Done.** |

**Milestone M2 (concluding month 18):**

| Delive-rable | Task | Responsible | Due mth. | Acc. | Description |
|---|---|---|---|---|---|
| D2.2 | T2.1 | LIMSI | 13 | R | Guidelines and testing protocols for the development of speech recognition components for SLDSs. |
| D2.3 | T2.2 | KTH | 13 | P | A survey of existing methods and tools for the development and evaluation of speech synthesis and speech synthesis quality in SLDSs. |
| D2.4 | T2.2 | KTH | 13 | R, S | Software tool for evaluation of speech synthesis components in SLDSs. |
| D2.5 | T2.3 | IMS | 13 | R | Survey of tools and methods for the acquisition of lexical data for SLDSs. Guidelines and tool specifications for checking, enhancing and extending the lexical coverage of SLDSs. This document will be discussed with the IAP. |
| D2.6 | T2.3 | IMS | 13 | R, S | Guidelines for the construction of linguistic resources for SLDSs. Guidelines for the representation of the relevant types of information. |
| D2.7 | T2.3 | MIP | 13 | P | State-of-the-art survey of existing dialogue management and human factors tools. |
| D2.8 | T2.4 | MIP | 13 | R, S | Concepts and a diagnostic methodology for the identification of user-system interaction problems, their typology, severity and remedies. Software tool in support of cooperative system dialogue design. |
| D2.9 | T2.4 | MIP | 13 | R, S | Software tool in support of speech functionality decisions in early design. |
| D3.9 | N/A | IAP | 15 | R | Assessment report on the DISC best practice methodology and toolbox. |
| D3.2 | T3.2 | LIMSI | 17 | R | Draft proposal on best practice methods and procedures in speech recognition. |
| D3.3 | T3.3 | KTH | 17 | R | Draft proposal on best practice methods and procedures in speech generation. |
| D3.4 | T3.4 | IMS | 17 | R | Draft proposal on best practice methods and procedures in language understanding and generation. |
| D3.5 | T3.5 | MIP | 17 | R | Draft proposal on best practice methods and procedures in dialogue management. |
| D3.6 | T3.6 | Vocalis | 17 | R | Draft proposal on best practice methods and procedures in human factors. |
| D3.7 | T3.7 | Vocalis | 17 | R | Draft proposal on best practice methods and procedures in systems integration. |
| MD3.8 | T3.8 | Vocalis | 18 | P | DISC Dialogue Engineering Best Practice Methodology. DISC software tools. |
| D4.4 | T4.2 | Elsnet | 18 | P | Best practice conference for industry and end-users. |
| MD5.2 | WP5 | MIP | 18 | P, A | Final report. |

**Figure 1.** Milestones and deliverables.

## 3.2 Deliverable Details Forms

**Project No.:** 24823 **Acronym:** DISC

DELIVERABLE DETAILS FORM
**Deliverable No.:** D1.1   **Due date:** 31.8.1998   **Date of finalisation:** 15.9.1998
**Short description:** The first DISC dialogue engineering best practice model.
**Partner responsible:** IMS.
**Partners who contributed:** IMS **with input from all partners**.
**Made available to:** Public.
**Description of further use, including exploitation of the deliverable results:**
**1. Inside the project:** Basis for work on WP1 in Year 1.
**2. Outside the project:** -
**Impact of the deliverable (publication, product, patent, contribution to standard, exhibition, technology transfer, etc.):** Used in publications on DISC in Year 1.

DELIVERABLE DETAILS FORM
**Deliverable No.:** D1.2   **Due date:** 31.3.1998   **Date of finalisation:** 16.5.1998
**Short description:** Working paper on speech recognition current practice. Based on exemplars analyses.
**Partner responsible:** LIMSI.
**Partners who contributed:** LIMSI, KTH.
**Made available to:** Restricted.
**Description of further use, including exploitation of the deliverable results:**
**1. Inside the project:** Basis for work on WP3 in Year 2.
**2. Outside the project:** Discussion with APs.
**Impact of the deliverable (publication, product, patent, contribution to standard, exhibition, technology transfer, etc.):** Publications, technology transfer.

DELIVERABLE DETAILS FORM
**Deliverable No.:** D1.3   **Due date:** 31.3.1998   **Date of finalisation:** 15.5.1998
**Short description:** Working paper on speech generation current practice. Based on exemplars analyses.
**Partner responsible:** KTH.
**Partners who contributed:** KTH, LIMSI.
**Made available to:** Restricted.
**Description of further use, including exploitation of the deliverable results:**
**1. Inside the project:** Basis for work on WP3 in Year 2.
**2. Outside the project:** Discussion with APs.
**Impact of the deliverable (publication, product, patent, contribution to standard, exhibition, technology transfer, etc.):** Publications, technology transfer.

DELIVERABLE DETAILS FORM
**Deliverable No.:** D1.4   **Due date:** 31.3.1998   **Date of finalisation:** 19.5.1998
**Short description:** Working paper on natural language understanding and generation current practice. Based on exemplars analyses.
**Partner responsible:** IMS.
**Partners who contributed:** IMS, LIMSI.
**Made available to:** Restricted.
**Description of further use, including exploitation of the deliverable results:**
**1. Inside the project:** Basis for work on WP3 in Year 2.
**2. Outside the project:** Discussion with APs.

**Impact of the deliverable (publication, product, patent, contribution to standard, exhibition, technology transfer, etc.):** Publications, technology transfer.

DELIVERABLE DETAILS FORM
**Deliverable No.:** D1.5    **Due date:** 31.3.1998   **Date of finalisation:** 25.5.1998
**Short description:** Working paper on dialogue management current practice. Based on exemplars analyses.
**Partner responsible:** MIP.
**Partners who contributed:** MIP, IMS, Linköping.
**Made available to:** Restricted.
**Description of further use, including exploitation of the deliverable results:**
**1. Inside the project:** Basis for work on WP3 in Year 2.
**2. Outside the project:** Discussion with APs.
**Impact of the deliverable (publication, product, patent, contribution to standard, exhibition, technology transfer, etc.):** Publications, technology transfer.

DELIVERABLE DETAILS FORM
**Deliverable No.:** D1.6    **Due date:** 31.3.1998   **Date of finalisation:** 18.5.1998
**Short description:** Working paper on human factors current practice. Based on exemplars analyses.
**Partner responsible:** Vocalis.
**Partners who contributed:** Vocalis, MIP.
**Made available to:** Restricted.
**Description of further use, including exploitation of the deliverable results:**
**1. Inside the project:** Basis for work on WP3 in Year 2.
**2. Outside the project:** Discussion with APs.
**Impact of the deliverable (publication, product, patent, contribution to standard, exhibition, technology transfer, etc.):** Publications, technology transfer.

DELIVERABLE DETAILS FORM
**Deliverable No.:** D1.7    **Due date:** 31.3.1998   **Date of finalisation:** 22.5.1998
**Short description:** Working paper on systems integration current practice. Based on exemplars analyses.
**Partner responsible:** Vocalis.
**Partners who contributed:** Vocalis, LIMSI, all.
**Made available to:** Restricted.
**Description of further use, including exploitation of the deliverable results:**
**1. Inside the project:** Basis for work on WP3 in Year 2.
**2. Outside the project:** Discussion with APs.
**Impact of the deliverable (publication, product, patent, contribution to standard, exhibition, technology transfer, etc.):** Publications, technology transfer.

DELIVERABLE DETAILS FORM
**Deliverable No.:** D1.8    **Due date:** 30.4.1998   **Date of finalisation:** 25.5.1998
**Short description:** Current practice in the development and evaluation of spoken language dialogue systems. Based on D1.1-D1.7.
**Partner responsible:** IMS.
**Partners who contributed:** IMS, MIP, all.
**Made available to:** Public.
**Description of further use, including exploitation of the deliverable results:**
**1. Inside the project:** Basis for work on WP3 in Year 2.
**2. Outside the project:** Discussion with APs.

**Impact of the deliverable (publication, product, patent, contribution to standard, exhibition, technology transfer, etc.):** Publications, technology transfer.

DELIVERABLE DETAILS FORM
**Deliverable No.:** D2.1   **Due date:** 30.6.1998 **Date of finalisation:** 15.5.1998
**Short description:** Survey of platforms and methods for speech recognition.
**Partner responsible:** LIMSI.
**Partners who contributed:** LIMSI.
**Made available to:** Public.
**Description of further use, including exploitation of the deliverable results:**
**1. Inside the project:** Basis for work on D2.2.
**2. Outside the project:** Discussion with APs.
**Impact of the deliverable (publication, product, patent, contribution to standard, exhibition, technology transfer, etc.):** Publications, technology transfer.

DELIVERABLE DETAILS FORM
**Deliverable No.:** D3.1   **Due date:** 31.5.1998   **Date of finalisation:** 1.6.1998
**Short description:** Working document on the detailed DISC best practice methodology.
**Partner responsible:** KTH.
**Partners who contributed:** KTH.
**Made available to:** Restricted.
**Description of further use, including exploitation of the deliverable results:**
**1. Inside the project:** Basis for work on WP3 in Year 2.
**2. Outside the project:** Discussion with APs.
**Impact of the deliverable (publication, product, patent, contribution to standard, exhibition, technology transfer, etc.):** Publications, technology transfer.

# 4. DISC-related Publications in Year 1

# 5. Aggregated information on resources used

|  | MIP | LIMSI | IMS | KTH | Vocalis | Daimler-Benz | Elsnet |
|---|---|---|---|---|---|---|---|
| **WP1** | | | | | | | |
| Planned PM | 8.25 | 7.5 +1 from KTH | 11 | 8.5 -1 to LIMSI | 9 | 0.25 | |
| PM used year 1 | 8 | 11 | 10.5 | 8 | 3 | 0.25 | |
| PM still to be used | 0.25 | 0 | 0.5 | 0 | 0 | 0 | |
| **WP2** | | | | | | | |
| Planned PM | 9 | 7 -1 to IMS | 7 +1 from LIMSI | 6 | | | |
| PM used year 1 | 5 | 5 | 3.25 | 4 | | | |
| PM still to be used | 4 | 3-4 | 4.75 | 2 | | | |
| **WP3** | | | | | | | |
| Planned PM | 6.75 | 6.5 | 6 | 6.5 | 9 | 0.75 | |
| PM used year 1 | 0.25 | 0 | 0.75 | 0.5 | 0 | 0 | |
| PM still to be used | 6.50 | 6.5 | 5.25 | 6 | 9 | 0.75 | |
| **WP4** | | | | | | | |
| Planned PM | | | | | | | 4 |
| PM used year 1 | | | | | | | 2.2 |
| PM still to be used | | | | | | | 1.8 |
| **WP5** | | | | | | | |
| Planned PM | 4.5 | | | | | | |
| PM used year 1 | 3 | | | | | | |
| PM still to be used | 1.5 | | | | | | |

**Notes:**

Vocalis: WP1: Staffing difficulties and the departure of David Williams at the End of March 1998 have resulted in the fewer hours costed to DISC than anticipated.

Elsnet: WP4: Planned manpower allocation for first year: 50 person days. Actually spent: 44 person days, mainly due to understaffing of the ELSNET office (Utrecht and Edinburgh) between November and February.
KTH: Gave 1 person month (PM) to IMS via LIMSI.
MIP: WP2: Delayed due to staffing shortage.

*******